\documentclass[lettersize,journal]{IEEEtran}
\usepackage{rotating}
\usepackage{amsmath,amsfonts}
\usepackage{algpseudocode}
\usepackage{array}
\usepackage{textcomp}
\usepackage{stfloats}
\usepackage{url}
\usepackage{verbatim}
\usepackage{relsize}
\usepackage{graphicx}
\usepackage{fontawesome}
\usepackage{cite}
\usepackage{hyperref}
\usepackage[table]{xcolor} 
\usepackage{array} 
\usepackage{graphicx} 
\usepackage{multirow}
\usepackage{amsmath,amssymb,amsfonts}
\usepackage{amsthm}
\usepackage{mathrsfs}
\usepackage{xcolor}
\usepackage{textcomp}
\usepackage{manyfoot}
\usepackage{booktabs}
\usepackage{color,soul}
\newcommand{\lightpinkhighlight}[1]{\sethlcolor{pink!30}\hl{#1}}
\newcommand{\lightbluehighlight}[1]{\sethlcolor{cyan!30}\hl{#1}}
\newcommand{\lightyellowhighlight}[1]{\sethlcolor{yellow!30}\hl{#1}}
\newcolumntype{C}[1]{>{\centering\arraybackslash}p{#1}}
\newcolumntype{C}[1]{>{\centering\arraybackslash}m{#1}}

\usepackage{algorithm}
\usepackage{algorithmicx}
\usepackage{algpseudocode}
\usepackage{listings}
\usepackage{amssymb}
\usepackage{tablefootnote}
\usepackage{booktabs}
\usepackage{array}
\usepackage{listings}
\usepackage{xcolor}
\usepackage{upquote} 
\usepackage{subcaption} 
\usepackage{float}
\usepackage{tabularx}
\usepackage{dirtytalk}
\usepackage[utf8]{inputenc}
\usepackage{csquotes}
\usepackage{array}
\usepackage{siunitx}
\usepackage{caption}
\begin{document}

\title{Unlocking Bias Detection: Leveraging
Transformer-Based Models for Content Analysis}
\author{Shaina Raza, 
Oluwanifemi Bamgbose,
Veronica Chatrath,
Shardul Ghuge,
Yan Sidyakin,
and Abdullah Y. Muaad
\thanks{S. Raza is with AI Engineering, Vector Institute, Toronto, ON, M5G 1M1, Canada, e-mail: (shaina.raza@vectorinstitute.ai). She is the corresponding author.}
\thanks{O. Bamgbose, V. Chatrath, S. Ghuge, and Y. Sidyakin are with the Vector Institute, Toronto, ON, M5G 1M1, Canada.}
\thanks{A. Y. Muaad is with the Department of Computer Science, University of Mysore, Mysuru, Karnataka, India.}
}



\maketitle

\begin{abstract}
Bias detection in text is crucial for combating the spread of negative stereotypes, misinformation, and biased decision-making. Traditional language models frequently face challenges in generalizing beyond their training data and are typically designed for a single task, often focusing on bias detection at the sentence level. To address this, we present the Contextualized Bi-Directional Dual Transformer (CBDT) \textcolor{green}{\faLeaf} classifier. This  model combines two complementary transformer networks: the Context Transformer and the Entity Transformer, with a focus on improving bias detection capabilities. We have prepared a dataset specifically for training these models to identify and locate biases in texts. Our evaluations across various datasets demonstrate CBDT \textcolor{green} effectiveness in distinguishing biased narratives from neutral ones and identifying specific biased terms. This work paves the way for applying the CBDT \textcolor{green} model in various linguistic and cultural contexts, enhancing its utility in bias detection efforts. We also make the annotated dataset available for research purposes. 
\end{abstract}


\section{Introduction}
\IEEEPARstart{A}{s} Natural Language Processing (NLP) rapidly evolves, its significance extends well beyond text analysis. NLP influences diverse sectors, ranging from social media analytics to advanced healthcare diagnostics. The pervasive reach of NLP showcases not only its achievements, but also highlights vital challenges. Among these challenges, linguistic biases \cite{raza2022dbias}, which are often embedded in both data and algorithms, are a significant concern. These biases do more than just perpetuate stereotypes; they risk distorting data interpretations, affecting decision-making processes.

\begin{figure}
    \centering
    \includegraphics[width=1\linewidth]{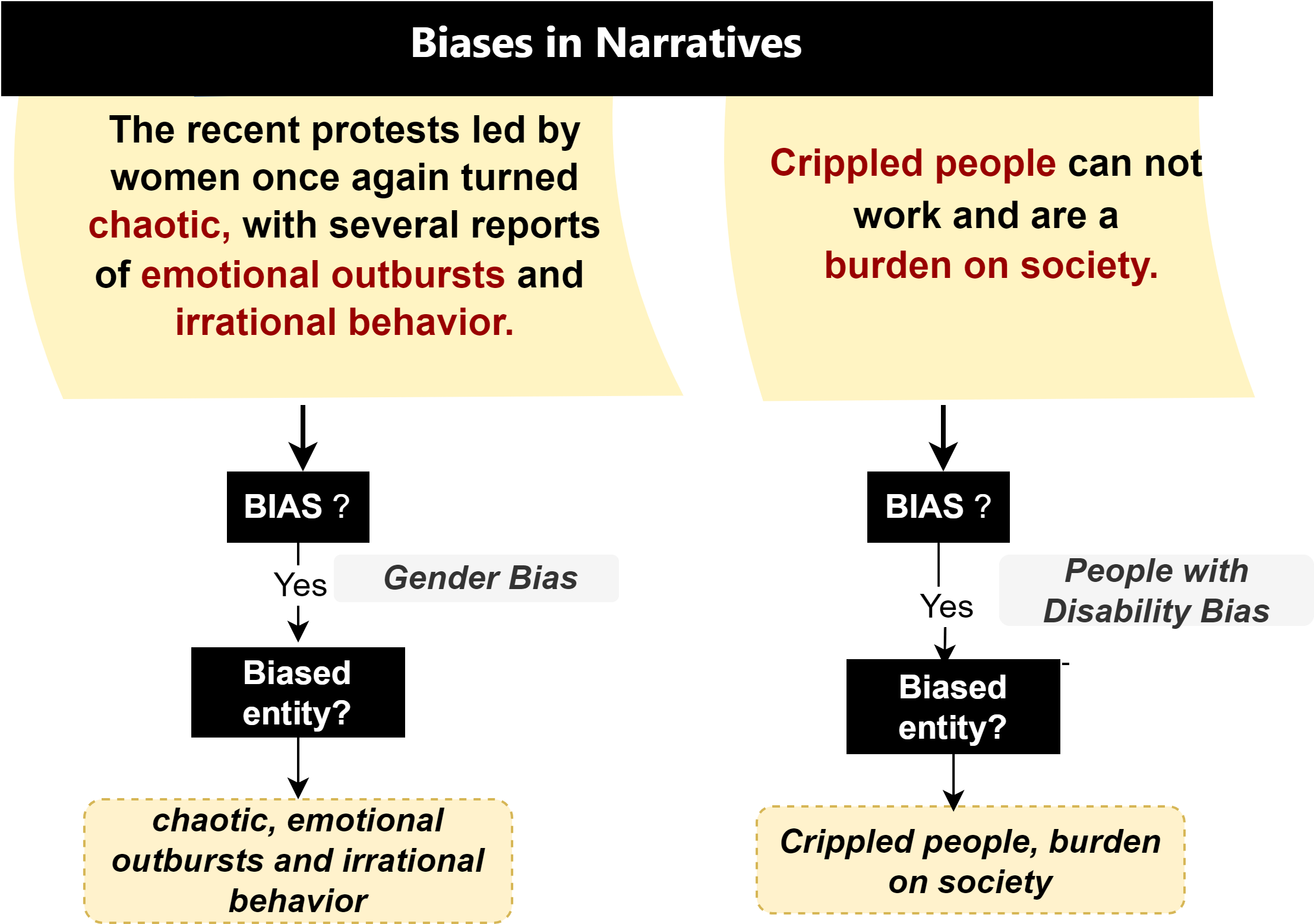}
    \caption{Real-world example of bias, highlighting the need for NLP solutions.}
    \label{fig:work}
\end{figure}

As seen in Figure \ref{fig:work}, the first statement displays a gender-based bias, while the second one reflects a bias for ableism. Here, \say{bias} refers to the predisposition or inclination towards a particular group (based on gender or race), often rooted in societal stereotypes, that can unduly influence the representation or treatment of particular groups \cite{Blondeletal2008}. These biases are often manifested in the tone, choice of words, or conveyed messages in textual conversations. A conventional NLP model primarily trained on a specific domain, such as political discourse, may detect bias within that domain but would be hesitant when presented with similar biases in a health-centric context. This inconsistency of traditional NLP models in effectively detecting bias across different domains highlights a gap in current NLP research: the capacity to generalize bias detection across diverse domains.

Despite significant advancements in state-of-the-art models~\cite{Liu2010,StereoSet,spinde2021neural,MeasuringSocialBiases,debiasingmethods,liu2023reducing}, consistent bias identification across diverse domains remains an ongoing challenge. Some recent research has mainly focused on evaluating and debiasing language models (LMs)~\cite{ungless-etal-2022-robust,joniak-aizawa-2022-gender,wang2023aligning} --- which is indeed a critical step toward minimizing AI risks --- the pressing need to detect biases inherent in the data itself, persists \cite{raza2024fair}. This highlights the urgency for a holistic NLP framework that not only stands up to academic observation, but also ensures ethical and accurate data interpretation.

To address this issue, our research introduces the novel \textit{Contextualized Bi-Directional Dual Transformer (CBDT)} \textcolor{green}{\faLeaf} classifier. At its core, the CBDT \textcolor{green}{\faLeaf} integrates two specialized transformer-based LMs: the Context Transformer and the Entity Transformer. The former assesses the overall bias in an input text, while the latter focuses on particular lexicons in the text to identify potentially biased words and entities. While many robust bias identification models \cite{zhou2021challenges, debiasingmethods, raza2022dbias, gallegos2023bias, raza2023nbias, lei-etal-2022-sentence, MeasuringSocialBiases, dev2021measures} are available in the literature, a significant portion of them possess a certain aspect of bias or scope, and their design is inherently singular. In contrast, we offer a dual-structured approach to accommodate a broader range of contexts and domains. Such an approach allows the CBDT \textcolor{green}{\faLeaf} classifier to comprehensively identify biases, offering a layered perspective by analyzing both overarching narrative structures and specific tokens within a text \footnote{\textcolor{red}{\textbf{\textcolor{red}{Warning: This paper contains content that may be perceived as offensive or biased, intended for research purposes.}}}}.

Our research offers the following key contributions:

\begin{enumerate}
    \item We introduce the CBDT \textcolor{green}{\faLeaf} classifier, a novel integration of two specialized transformer architectures: the first transformer performs contextual understanding of biases at the sentence level, and the second is aimed at entity-level bias identification. Identifying bias narratives at the sentence level is a common task, but combining it with token-level bias identification is a unique approach. In particular, CBDT \textcolor{green}{\faLeaf} is one of the first to customize bias identification at the token level. 
    \item Recognizing the importance of data quality, we put forth a carefully curated corpus that covers both overt and subtle biases. Drawing from domain expertise, lexicon creation, and rule-based formulation, our training dataset emerges as a diverse, comprehensive, and representative collection. We have made our data available as a contribution to the research community.

    \item We extend our model evaluation beyond our primary dataset to include evaluation on out-of-distribution datasets. This thorough approach underscores our model's robustness, flexibility, and consistent performance across a plethora of contexts. We also demonstrate the performance of the CBDT \textcolor{green}{\faLeaf} classifier using various quantifiable measures and show that its performance achieves both determining if a narrative is biased or not and pinpointing bias lexicons within the texts.
\end{enumerate}

\section{Related Work}
\noindent NLP systems have long been susceptible to the issue of bias, leading to unfair representation in their outcomes \cite{garrido2021survey}. Numerous studies \cite{caliskan2017semantics, spinde2021neural,raza2022fake} have highlighted how societal and cultural biases inadvertently enter training data. These biases can undermine the integrity of NLP outcomes, perpetuating, and at times amplifying, societal disparities \cite{raza2022dbias}.

The study of bias addresses various forms of discrimination and misrepresentation in text \cite{mendelson_debiasing_2021}. A wealth of research has been directed at understanding and mitigating biases in language models and embeddings \cite{raza2022fake, gallegos2023bias,ding2022word}. Significant findings include the identification of gender bias in popular embeddings like GloVe and Word2Vec \cite{bolukbasi2016man}, and the development of the Word Embedding Association Test (WEAT) to quantify biases in word embeddings \cite{caliskan2017semantics}. This methodology was later extended to assess biases in sentence encoders \cite{MeasuringSocialBiases}. There have been documented efforts to reduce biases in BERT by fine-tuning with counterfactual datasets \cite{bartl2020unmasking}. Research has also delved into gender and representation biases in GPT-3 \cite{lucy2021gender}, and the perpetuation of biases in conversational AI systems \cite{barikeri_redditbias_2021}, highlighting the critical importance of bias mitigation in dialogue generation. These efforts collectively contribute to the pursuit of more equitable and unbiased language technologies.

In a related work, a system has been asked to detect hate speech and provide explanations \cite{cai2022power}. Concurrently, another study explored biases in text-based event detection, addressing both data scarcity and annotation challenges \cite{wang2023m4}. The research presented in \cite{cheng2022toward} investigates the relations between different forms of biases in NLP models, specifically examining bias mitigation in toxicity detection and word embeddings. This study concentrates on three social identities: race, gender, and religion, suggesting that biases can be correlated and that standalone debiasing methods may prove inadequate. 

There is a rich diversity in research focusing on different facets of bias in NLP. Govindarajan \textit{et al.} \cite{govindarajan2023people}, for instance, shift the focus from demographic biases to predict group dynamics using emotional cues. In another work, a critical analysis \cite{devinney2022theories} was conducted on gender biases in NLP studies, stressing the lack of gender-focused theories. A novel methodology \cite{zhao2023combating} was introduced to counter dataset biases employing a gradient alignment strategy. These insights emphasize the need for continuous vigilance and proactive measures to ensure fairness in NLP models.

Bauer \textit{et al.} \cite{bauer2021analyzing} define and extensively evaluate how well language models grasp the semantics of four bias-related tasks: diagnosis, identification, extraction, and rephrasing. This evaluation reveals that LMs can handle these tasks to varying extents across multiple bias dimensions, including gender and political orientation. Additional research \cite{zhou2023causal} introduces a cohesive framework that adeptly minimizes unwanted biases in LMs during fine-tuning for subsequent tasks without compromising performance. 

Abid \textit{et al.} identified persistent Muslim-violence bias in GPT-3.5 \cite{abid2021persistent}. A related method \cite{lei-etal-2022-sentence} was proposed to identify biases at the sentence level within news articles. The work demonstrated that understanding the discourse role of a sentence and its relation with nearby sentences can reveal the ideological leanings of an author. Fu \textit{et al.} \cite{fu2023gender} utilize NLP techniques to detect language bias in letters of recommendation. The research employed methods such as sublanguage analysis, dictionary-based approach, rule-based approach, and deep learning approach to extract psycholinguistics and thematic characteristics in letters.

An interdisciplinary approach was advocated to understand bias in NLP \cite{narayanan2023towards}. The research emphasized identifying sociodemographic bias in various NLP architectures and the importance of interdisciplinary collaboration to bridge gaps and foster a more inclusive and accurate analysis of bias. A comprehensive taxonomy was developed by Doughman \textit{et al.} \cite{doughman2021gender} to help identify and mitigate relevant biases from training corpora for improved fairness in NLP systems.

Additionally, Garrido \textit{et al.} \cite{garrido2021survey} formally introduced the concept of bias in deep NLP, suggesting methods for its detection and rectification. Gallegos \textit{et al.} \cite{gallegos2023bias} delved deeper into the concepts of social bias and fairness in NLP, with the goal of actualizing fairness in LMs. These studies underline the significance of recognizing and addressing biases in the ever-advancing domain of NLP.

While the methods previously mentioned have greatly influenced our research, our work stands distinct. Unlike prior studies, we have devised a strategy for corpus construction specifically tailored to detect biases in texts. Additionally, we employ a dual LMs approach, enabling bias identification both at the sentence and narrative levels.

\section{Contextual Bi-Directional Transformer (CBDT {\color{green}\faLeaf}~) Framework}
\label{sec:method}

\begin{figure*}
    \centering
    \includegraphics[width=.8\linewidth]{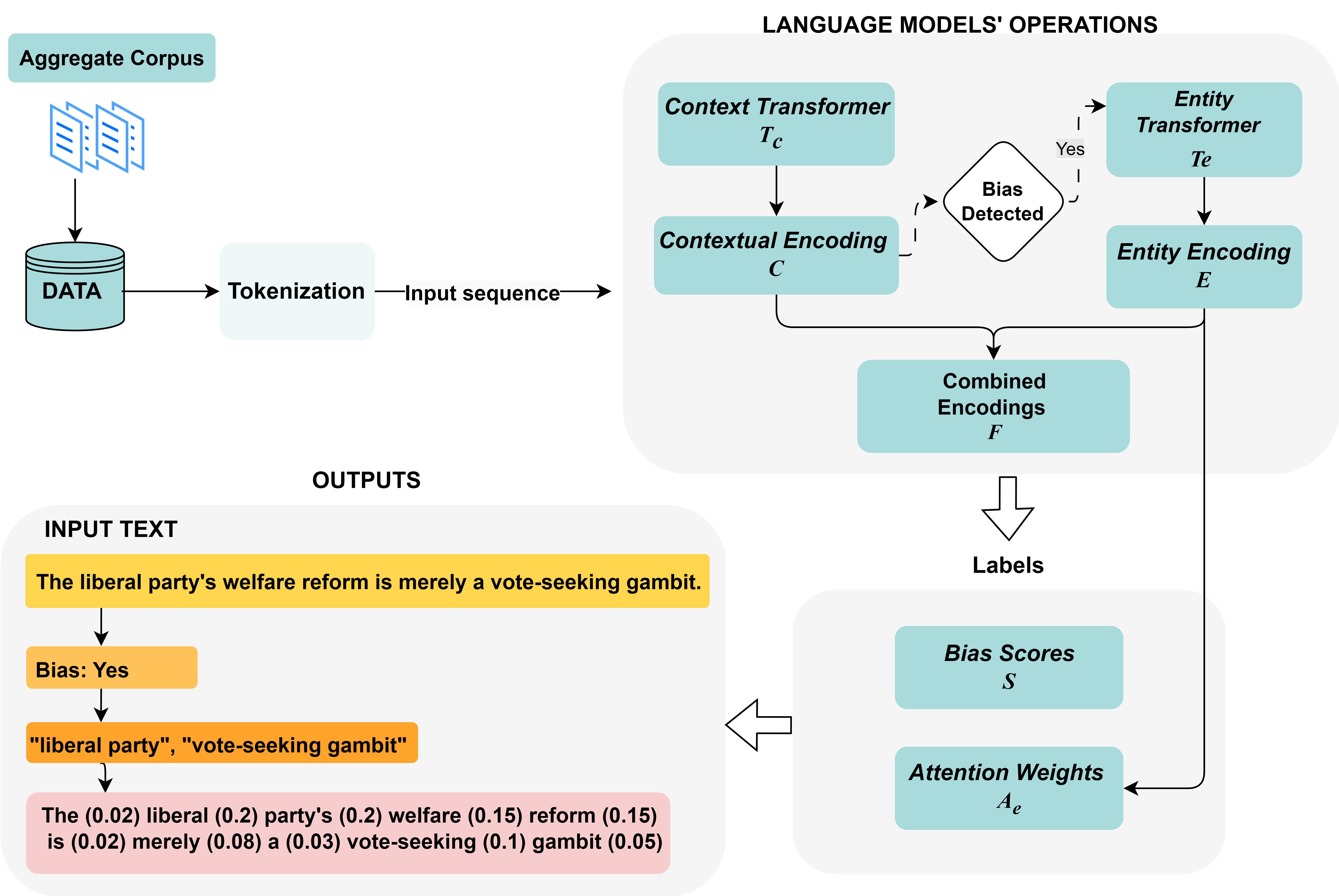}
    \caption{A visual representation of the CBDT  {\color{green}\faLeaf}  method's workflow for bias detection in textual data.}
    \label{fig:archi}
\end{figure*}

\noindent Figure \ref{fig:archi} depicts our CBDT {\color{green}\faLeaf} architecture. The architecture consists of two primary components: a corpus construction and a dual-transformer structure. The latter combines the Context Transformer, which shows the contextual understanding of biases at the sentence level, and the Entity Transformer, aimed at identifying biased tokens as entity. Both transformers are seamlessly integrated into the language model's operations. The resulting output displays whether a sentence is biased or not, along with specific words or entities (tokens) that contribute to the detected bias.

\subsection{Corpus Construction}
\label{sec:corpus}

\noindent The corpus construction steps are given below:
\paragraph{Defining Bias Dimension} A bias dimension represents the overarching theme or category of bias. It is grounded in societal, cultural, or stereotypical perspectives leading to preferential or prejudicial views toward specific groups or individuals \cite{doughman2021gender,devinney2022theories}. Recognizing these dimensions is crucial for accurate identification and analysis of bias in various textual contexts. 

\paragraph{Lexicon Creation and Rule Formulation} To address this need, we assembled a multi-disciplinary panel comprising two experts from computational linguistics and two domain specialists from journalism and health sectors. Their collective expertise guarantees a holistic capture of potential biases originating from various dimensions. This panel carefully curated a lexicons list populated with biased terms, phrases, and structures, specifically focusing on expressions indicative of stereotypes or manifesting explicit biases. A concise version of these lexicons is showcased below:

\begin{itemize}
    \item \textbf{Race} - illegal alien, thug, savage, barbaric
    \item \textbf{Gender} - emotional, weak, bossy, hysterical
    \item \textbf{Religion} - radical, terrorist, fanatic, heretic
    \item \textbf{Age} - slow, old-fashioned, senile, out of touch
    \item \textbf{Sexual Orientation} - unnatural, abnormal, deviant, sinful
    \item \textbf{Profession} - greedy, dishonest, shady, untrustworthy
    \item \textbf{Social Status} - lazy, freeloader, bum, worthless
    \item \textbf{National} - unpatriotic, alien, foreigner, outsider
    \item \textbf{Disability} - crippled, handicapped, defective, impaired
    \item \textbf{Education} - uneducated, illiterate, backward, naive
    \item \textbf{Body Size} - fat, slob, whale, blob
    \item \textbf{Climate} - hoax, alarmist, tree-hugger, denier
    \item \textbf{Political} - extremist, radical left, far-right, ideologue
    \item \textbf{Economic Status} - welfare queen, gold-digger, elitist, tycoon
    \item \textbf{Region} - hillbilly, city slicker, rust belter, coastal elite
    \item \textbf{Ethnicity} - exotic, oriental, primitive, tribal
    \item \textbf{Cultural} - uncultured, barbarian, savage, civilized
    \item \textbf{Lifestyle} - hipster, jock, nerd, geek
    \item \textbf{Appearance} - frumpy, plain, flashy, ostentatious
    \item \textbf{Health/ Wellness Narrative} - fragile, feeble, unfit, sturdy
\end{itemize}

We have introduced a range of bias dimensions accessible for use in our dataset. Additionally, experts have gathered and curated approximately 2000 rules for identifying bias. These rules comprise a sentence, its associated aspect/dimension of bias, and the rationale behind its bias. The objective is to employ these rules to develop annotation guidelines.

\paragraph{Embedding and Bias Detection} Using BERT, we generate embeddings for every sentence in the dataset in a semi-supervised manner. When the cosine similarity between the embeddings of a predefined rule and a text surpassed a specific threshold, the text was flagged as potentially biased. The matching rule then determines its bias dimension. For increased accuracy, we employ an exact word matching method against the bias lexicon. Sentences with matching words are flagged as potentially biased.

\paragraph{Bias Label Assignment} The \say{bias\_label} is assigned based on a combination of lexicon matching, rule-based flagging, and BERT-based similarity measures. Sentences flagged by any of these methods are marked for manual review by domain experts.

All flagged content undergoes a manual review by four domain experts. Cohen’s Kappa and Fleiss’ Kappa are used to measure the agreement among annotators, with our rigorous validation process yielding high values of 0.75 and 0.72, respectively, showing a substantial agreement among annotators.

\paragraph{FAIR Scheme Adoption} Our dataset adopts the FAIR principles — Findable, Accessible, Interoperable, and Reproducible — as highlighted in \cite{raza2024fair}. This scheme ensures greater accessibility and usability for research purposes. By assigning unique identifiers and providing detailed metadata, we make the dataset easily \lightbluehighlight{findable} for researchers and automated systems. It is hosted on a stable, accessible platform, with explicit instructions facilitating straightforward \lightbluehighlight{accessibility}. Compliance with widely accepted standards, such as the CoNLL-2003 format, guarantees \lightbluehighlight{interoperability} with a variety of tools and research environments. Furthermore, comprehensive documentation, including ethical guidelines and licensing information, underpins the dataset's \lightbluehighlight{reusability} across different contexts.

To underscore the effectiveness of our methodology in identifying biased language within texts, we have adopted a format that is notably enhanced by our use of the `Binary Label Format' for classification labels. This format simplifies the determination of the presence or absence of bias. The specific bias aspects considered in our data along with a binary representation for easy identification of biased content, can be accessed at the provided link\footnote{\url{https://huggingface.co/datasets/newsmediabias/bias-aspects}}. Additionally, we adhere to the CoNLL-2003 format \cite{sang2003introduction} for identifying bias tokens, making the data available here\footnote{\url{https://huggingface.co/datasets/newsmediabias/Bias-Tokens-CONLL}}. Moreover, we have meticulously outlined the ethical guidelines, licensing, and usage policies of our dataset in \cite{raza2024fair}, ensuring transparency and ethical compliance in our research practices. This comprehensive approach not only facilitates precise bias detection but also adheres to the highest standards of research integrity and data utilization.

The pseudocode for this approach is in Algorithm \ref{algo:corpus}.

\begin{algorithm}
\caption{Concise Corpus Preparation and Labelling}
\label{algo:corpus}
\begin{algorithmic}[1]
\State \textbf{Initialize} datasets and lexicon

\Function{PreprocessAndLexicon}{datasets}
    \State Clean data, removing special characters, lowercase, handle missing values.
    \State Assemble expert panel, discuss inclusion criteria, and curate biased terms.
    \State \Return consolidated\_dataset, lexicon
\EndFunction

\Function{DetectAndLabelBias}{consolidated\_dataset, lexicon}
    \For{each sentence in consolidated\_dataset}
        \State Use BERT for embeddings, determine threshold, flag bias.
        \If{matches lexicon OR BERT similarity}
            \State Assign \say{bias\_label}, flag for review.
        \EndIf
    \EndFor
    \State Discuss discrepancies among reviewers; reach consensus.
    \State \Return sentences with bias labels
\EndFunction

\Function{FinalizeDataset}{labeled\_sentences}
    \State Annotate with metadata, format according to FAIR, host dataset.
    \State \Return FAIR\_conformant\_dataset
\EndFunction

\State consolidated\_dataset, lexicon $\gets$ \Call{PreprocessAndLexicon}{datasets}
\State labeled\_sentences $\gets$ \Call{DetectAndLabelBias}{consolidated\_dataset, lexicon}
\State output $\gets$ \Call{FinalizeDataset}{labeled\_sentences}

\State \textbf{Output} output
\end{algorithmic}
\end{algorithm}

\textbf{Dataset Schema:} Our dataset is designed to align with binary classification format and CoNLL-2003 standards. It encompasses a variety of sources, such as the BABE dataset \cite{spinde2021neural} which focuses on political and news media biases. For detecting biases within health notes related to specific genders or races, we incorporated MIMIC-IV clinical notes \cite{johnson2020mimic} \footnote{Due to licensing restrictions, the MIMIC Dataset notes are not made available in the data release.} Additionally, we compiled data on climate change news and occupational reports using Google RSS feeds, with a coverage period ranging from January 2023 to July 2023. More details of these datasets are in Table \ref{datasets}.

\noindent In the binary format, the sentence is paired with a binary score indicating the presence or absence of bias. The CoNLL-2003 format, on the other hand, is more granular, tagging each word/token in the sentence to signify whether it is part of a biased expression \cite{Raza2023}. 

\begin{lstlisting}[caption={Schema for Bias Annotation}]
{
  biased_text: {
    Type: 'String',
    Example: 'A certain group is 
    always lazy...'
  },
  bias_label: {
    Type: 'Boolean',
    Example: 'true'
  },
  identified_biased_spans: {
    Type: 'List of Strings',
    Example: ['always lazy']
  },
  bias_dimension: {
    Type: 'String',
    Example: 'Ethnic Stereotyping'
  }
}
\end{lstlisting}

\begin{lstlisting}[caption={CoNLL-2003 Format Example}]
            A       O
            certain O
            group   O
            is      O
            always  B-BIAS
            lazy    I-BIAS
            .       O
\end{lstlisting}

Prior to formal experimentation, we select representative excerpts from each domain, showcasing potential bias indicators like emotive language, stereotypical portrayals, or blame attribution. These extracts are presented in Appendix A.

\subsection{Method}

\noindent The Contextual Bi-Directional Transformer (CBDT {\color{green}\faLeaf}) method introduces a novel approach to bias detection in textual data. This methodology operates in two main stages: the Context Transformer ($T_c$) for sentence-level bias detection, and the Entity Transformer ($T_e$) for token-level bias identification.

Given a tokenized input text, $X = \{x_1, x_2, \ldots, x_n\}$, the $T_c$ first assesses the entire sequence to determine the presence of bias, generating a contextual encoding, $C$. This encoding encapsulates the sentence-level bias context.

If bias is detected, $T_e$ further analyzes the sequence to pinpoint specific tokens or entities contributing to the bias, resulting in an entity encoding, $E$. The outputs of both transformers, $C$ and $E$, are concatenated to form a composite feature representation, $F = [C; E]$, which is then fed into a classifier to determine a bias score, $S$. The score, $S$, could be a continuous value indicating the level of bias, with a specific range for interpretation.

Additionally, $T_e$ produces a set of attention weights, $A_e = \{a_1, a_2, \ldots, a_n\}$, where each weight, $a_i$, reflects the relative contribution of token $x_i$ to the overall bias. These attention weights offer insights into which tokens are most influential in the perceived bias, aiding interpretability.

Both the Context and Entity Transformers are based on BERT uncased models, with $T_c$ functioning as a binary classifier for initial bias detection and $T_e$ identifying specific biased tokens.

Evaluation of the CBDT {\color{green}\faLeaf} model's effectiveness in bias detection can be done using metrics such as accuracy, precision, recall, and F1-score for sentence-level classification, and token-level or span-level F1-scores for token classification tasks. These metrics will help quantify the model's performance in accurately detecting and highlighting biased text.

\begin{algorithm}
\caption{CBDT  {\color{green}\faLeaf}  Bias Detection and Training}
\label{alg:CBDTdetection}
\begin{algorithmic}[1]
\Procedure{CBDT  {\color{green}\faLeaf} }{$X$}
    \State \textbf{Input}: Sequence of tokens $X = \{x_1, x_2, \ldots, x_n\}$
    \State \textbf{Output}: Bias score $S$, Attention weights $A_e$
    
    \State $C \gets T_c(X)$ \Comment{Context Transformer detects bias as a score or flag}
    
    \If{$C$ indicates bias (e.g., score $>$ threshold)}
        \State $E \gets T_e(X)$ \Comment{Entity Transformer for biased tokens}
        \State $A_e \gets$ Extract attention weights from $T_e$
    \Else
        \State $E \gets$ Zero vector matching expected dimensions \Comment{No bias detected}
        \State $A_e \gets$ Zero vector matching number of tokens in $X$
    \EndIf
    
    \State $F \gets [C; E]$ \Comment{Merge contextual and entity encodings}
    \State \textbf{return} Classifier($F$), $A_e$ \Comment{Bias score and token attention weights}
\EndProcedure

\Procedure{TrainCBDT  {\color{green}\faLeaf} }{$D$, epochs, mini\_batch\_size}
    \State \textbf{Input}: Labeled dataset $D = \{(X_1, y_1), \ldots, (X_m, y_m)\}$
    \State \textbf{Output}: Optimized model parameters $\theta$
    
    \For {epoch in 1, \ldots, epochs}
        \For {each mini-batch of $(X_i, y_i)$ in $D$}
            \State $S, A_e \gets$ \Call{CBDT  {\color{green}\faLeaf} }{$X_i$}
            \State Compute loss $L(y_i, S)$ using binary cross-entropy for binary $y_i$
            \State Update $\theta$ using Adam optimizer with specified learning rate
        \EndFor
    \EndFor
    \State \textbf{return} Optimized $\theta$
\EndProcedure
\end{algorithmic}
\end{algorithm}

\section{Experimental Design}
\label{sec:expDesign}
The experimental design for the study is given as: 

\subsection{Training and Evaluation Data}
\label{eval-data}
\noindent The specifics of the training dataset are found in Table \ref{datasets}.  
 For our evaluations, we used a test set from our combined data, split as 70-15-15,  Additionally, two other out-of-distribution test sets are used in measuring the models' bias-detection capabilities:
\begin{itemize}
    \item \textit{SemEval-2018 Valence Classification Data}  \cite{mohammad-etal-2018-semeval}:  Used for sequence classification, this dataset was adapted for binary bias detection. We specifically focused on the English test set with 3259 records.
    \item \textit{CoNLL 2003 Shared Task Data }\cite{sang2003introduction}: Applied for token classification, we utilized the testset that covers English content with 231 articles, with 3684 sentences, and 46435 tokens. The BIO labeling scheme facilitated token-based categorization \cite{ratinov2009design}.
\end{itemize}

\begin{table}[htbp]
\centering
\caption{Data sources, domains, and split for Training, Development, and Testing. \tablefootnote{Annotated data is made available for research.}}
\scalebox{0.95}{
\begin{tabular}{lcccc}
\toprule
\textbf{Data Source} & \textbf{Domain} & \textbf{Train} & \textbf{Dev} & \textbf{Test} \\
\midrule
BABE \cite{spinde2021neural} & News Media & 2590 & 555 & 555 \\
Google RSS & Climate Change & 1050 & 225 & 225 \\
Google RSS & Occupations & 1500 & 225 & 225 \\
MIMIC-IV \cite{johnson2020mimic} & Clinical Notes & 1400 & 300 & 300 \\
SemEval-2018 (en.)\tablefootnote{\say{\textit{en}} means \say{\textit{English}}.} \cite{mohammad-etal-2018-semeval} & Valence Classification & -- & -- & 3259 \\
CoNLL 2003 (en.) \cite{sang2003introduction} & Shared Task Data & -- & -- & 3684\\
\bottomrule
\end{tabular}}
\label{datasets}
\end{table}

The training dataset has been aggregated, incorporating a total of 6540 training samples. This consolidation was undertaken to enhance the diversity and comprehensiveness of the dataset, crucial for improving model robustness and accuracy. This approach ensures that the training process benefits from a rich, varied set of examples, thereby enhancing the model's ability to generalize from the training data to new, unseen data.

\subsection{Evaluation Strategy}
\noindent Our approach is systematically compared against a diverse set of baselines across three main categories: \lightyellowhighlight{traditional algorithms} such as Naïve Bayes (NB), Support Vector Machine (SVM), and RandomForest (RF). While RF, SVM, and NB are not inherently designed as sequence classification models, we adapt them for this purpose to provide a comprehensive comparison. For example, by converting sequences into a suitable vector representation (e.g., using TF-IDF for text, embedding vectors, and feature extraction methods where applicable) to capture sequence information, these models can be effectively used for sequence classification.

We also use \lightyellowhighlight{neural network-based approaches} including Convolutional Neural Network (CNN) and Long Short-Term Memory (LSTM); and advanced models like  \lightyellowhighlight{state-of-the-art large language models (LLMs)}, such as DeBERTa \cite{DeBERTa}, RoBERTa \cite{liu2019roberta}, GPT-2 \cite{radford2019language}, GPT-3.5 (GPT-3.5-turbo) \cite{gpt3}, and Falcon 7B \cite{falcon40b}. Specifically, GPT-2 is fine-tuned with a classification layer, GPT-3.5 is applied in a few-shot setting, and Falcon 7B operates in a zero-shot configuration.

The fine-tuned baselines can be divided into two primary groups:  \lightyellowhighlight{probability-based baselines}, which are assessed on their capability to determine the likelihood of content being biased or non-biased, and  \lightyellowhighlight{prompt-based baselines}, which employ prompts in zero-shot or few-shot settings for evaluation. This classification scheme draws inspiration from Llama Guard \cite{inan2023llama}, aiming to categorize distinct types of evaluations effectively.

We employ two primary kinds of metrics to assess the bias detection capabilities of our model alongside these baselines. For \textit{probability-based baselines}, models are evaluated on their ability to classify content as biased or non-biased, using accuracy, precision, recall, F1-score, and AUC. The token-classification performance is further examined using the macro-averaged F1-score, catering to multi-label classification scenarios \cite{nadeau2007survey}. These metrics collectively offer a comprehensive assessment of a model's precision in identifying bias, its efficiency in minimizing false positives, and its overall discriminatory power.

For \textit{prompt-based baselines} relevant to LLMs like GPT-3.5 and Falcon-instruct, we utilize the GPT-score \cite{fu_gptscore_2023} model to derive confidence scores for classifying content, with a threshold set above 0.5. This evaluation framework, particularly through the application of GPT-3.5 for result scoring, provides a nuanced understanding of the model's proficiency in bias detection under instruction-based conditions.

This dual evaluation strategy ensures a thorough and multi-dimensional analysis of model performance, highlighting the strengths and potential areas for improvement in detecting biases across various textual contents.

\subsection{Settings}
The fine-tuning of the model was conducted on a high-performance computing setup, using a single NVIDIA A40 GPU, supported by 4 CPU cores. This hardware configuration was chosen for its balance of computational power and efficiency, enabling rapid model iterations and extensive experimentation. The software environment was standardized across all experiments, employing TensorFlow 2.x and PyTorch 1.x for model development and training. These libraries were selected for their extensive support for deep learning models and compatibility with the NVIDIA A40 GPU. This setup facilitated the fine-tuning of the models, allowing for precise adjustments to model parameters and optimization strategies to enhance bias detection capabilities in textual data.

\textbf{Carbon Footprint} Given the energy-intensive nature of training transformer-based models, estimating the carbon footprint \cite{dodge_measuring_2022} is crucial for understanding the environmental impact. For our CBDT {\color{green}\faLeaf}, trained on a single NVIDIA A40 GPU for 4 epochs with each epoch taking about 10 minutes, the calculation proceeds as follows:
The NVIDIA A40 GPU has a maximum power consumption of 300 watts. Since the GPU operates at full capacity during training, the energy consumption per epoch is calculated as $0.3 \, \text{kW} \times 10 \, \text{min} = 0.05 \, \text{kWh}$. Over 4 epochs, this amounts to $0.2 \, \text{kWh}$.  
Considering the global average carbon intensity of electricity generation is roughly $0.5 \, \text{kgCO}_2\text{e}/\text{kWh}$, the total carbon emissions for training our model for one complete cycle are $0.2 \, \text{kWh} \times 0.5 \, \text{kgCO}_2\text{e}/\text{kWh} = 0.1 \, \text{kgCO}_2\text{e}$.

 \textbf{Hyperparameters} \noindent For the CBDT  {\color{green}\faLeaf}  classifier, we use a learning rate of 0.001 with Adam, a batch size of 64, 20 epochs, the ReLU and Softmax activation functions, a dropout of 0.5, and a weight decay of 0.0001. All the baseline methods are fine-tuned to their best hyperparamter settings.  CBDT  {\color{green}\faLeaf}  is fine-tuned for specific tasks, while Naïve Bayes uses TF-IDF for text vectorization. SVM employs a linear kernel, and RandomForest uses 100 estimators. CNN relies on 1D convolutional layers, and LSTM utilizes 128 LSTM units. DeBERTa and RoBERTa are used with the base-uncased version and are fine-tuned. GPT-2 is fine-tuned, adding a classification layer on top, whereas GPT-3.5 is API-based with a temperature setting of 0.7 for response generation.  For Falcon 7B, we use the ``falcon-7b-instruct" model and the model has been quantized using the BitsAndBytes \cite{belkada_younes_making_2023} library to load parameters and activations in 4-bit format, enabling more efficient computation. More hyper parameter details are in Appendix D. 

\section{Results}\label{sec:results}
\noindent In our results section, we assess the CBDT {\color{green}\faLeaf} model's efficacy in handling two interrelated tasks: sequence classification and token classification. The Context Transformer initially evaluates whether a sequence contains bias. Upon detecting bias, the Entity Transformer further investigates to identify the specific tokens or entities exhibiting bias. We examine the effectiveness of both transformers in these evaluations.
 We also conduct an ablation study to underscore the significance of each module within the CBDT  {\color{green}\faLeaf} architecture, illustrating the model's robustness in diverse settings. 

\subsection{Evaluating Context Transformer for Sequence Classification Task}

\begin{table}[h]
\centering
\caption{Comparative Performance of various classification models on two test sets. \textbf{Top:} Our Test Set and \textbf{bottom:} SemEval. Higher scores are better and highlighted in \textbf{bold}. All models are fine-tuned, except for Falcon7B and GPT3.5, which are tested on zero-shot (ZS) and few-shot (FS) setting, respectively. RF stands for Random Forest, SVM stands for Support Vector Machine, and NB stands for Naïve Bayes.
}
     \begin{tabular}{llccc}
    \toprule
    & Model & Precision (\%) & Recall (\%) & F1-Score (\%) \\
  \midrule
\parbox[t]{1mm}{\multirow{10}{*}{\rotatebox[origin=c]{90}{Test Set}}} & RF& 80.48& 81.20& 80.84\\
& SVM & 82.30& 82.79& 82.54
\\
& NB & 80.48& 81.20& 80.84
\\
& LSTM & 81.78& 81.53& 81.65
\\
 & CNN& 85.45& 82.67&84.04\\
& DeBERTa & 88.11& 83.41& 85.69
\\
& RoBERTa & 89.10& 88.94& 90.40
\\
& Falcon7B (ZS) & 84.42& 82.18& 83.28\\
 & GPT-2 (FT) & 58.18& 69.89&63.50
\\
& GPT-3.5 (FS) & 88.41& 89.28& 88.84
\\

 \cmidrule{2-5}
& CBDT  {\color{green}\faLeaf} (ours) & \textbf{93.44}& \textbf{94.30}& \textbf{93.87}
\\
\midrule

\parbox[t]{1mm}{\multirow{10}{*}{\rotatebox[origin=c]{90}{SemEval}}} & RF & 74.57 & 76.02 & 75.31 \\
& SVM & 75.60 & 77.36 & 76.45 \\
& NB & 76.49 & 78.94 & 77.77 \\
& LSTM & 78.04 & 80.79 & 79.42 \\
& CNN & 83.71 & 82.47 & 83.01 \\
& DeBERTa & 84.40 & 83.35 & 83.90 \\
& RoBERTa & 83.41 & 86.16 & 84.63 \\
& Falcon7B (ZS) & 79.23 & 81.98 & 80.61 \\
& GPT-2 (FT) & 59.23 & 58.44 & 58.83 \\
& GPT-3.5 (FS) & 85.00 & 87.75 & 86.38 \\
 \cmidrule{2-5}
& CBDT  {\color{green}\faLeaf} (ours) & \textbf{89.25} & \textbf{91.85} & \textbf{90.48} \\
\bottomrule
\end{tabular}
\label{tab:sequnce2}
\end{table}

\noindent Table \ref{tab:sequnce2} presents the comparative performance of various models for the sequence classification task.

The CBDT {\color{green}\faLeaf} model, which is our main contribution, consistently outperforms all the baseline models across both evaluation sets. Specifically, on the primary test set, it achieves an F1-score of  93.44\%. On the SemEval dataset, it achieves an F1-score of  90.48\%. These results underscore the CBDT {\color{green}\faLeaf} model's effectiveness in addressing nuanced bias aspects. 

Traditional algorithms like RF, SVM, and NB have shown competitive results. The F1-scores for these models oscillate between 74\% to 83\% across both testsets. However, when compared against most of the  network-based architectures like BERT-based models and GPT-3.5, their limitations become evident.

Within the domain of neural network-based models, LSTM and CNN showed commendable performance. Nevertheless, their performance is lower (and in some cases marginally below) than  transformer-based models such as GPT-3.5, DeBERTa, RoBERTa, and CBDT  {\color{green}\faLeaf} . This underscores the potential and superiority of transformer-based LM architectures, especially in bias detection tasks. Although GPT-3.5 few-shot learning yields commendable outcomes, it performs less than a fine-tuned model like CBDT  {\color{green}\faLeaf}. The zero-shot learning approach of Falcon 7B also exhibits decent performance, but it is not one of the best-performing. This shows that while zero-shot learning has its merits, fine-tuning or employing task-specific models like CBDT {\color{green}\faLeaf} can elevate the performance ceiling.

A noticeable decrease in performance is observed across all models when transitioning from the primary test set to the SemEval dataset. This highlights the inherent challenges posed by out-of-distribution data. However, the decline in performance for the CBDT {\color{green}\faLeaf} model is relatively more restrained, which hints at its superior generalization capabilities.  
We also show the confusion matrices for different models for the sequence classification task in Appendix C. These confusion matrices provide insights into the classification performance of each model, allowing for further analysis and comparison of their efficacy in the sequence classification task. Overall, we observe that best-performing models like CBDT  {\color{green}\faLeaf} tend to have lower false positive and false negative rates.
 
\paragraph{Models' Performance Over Training Epochs}
\begin{figure}[h]
     \centering
     \includegraphics[width=1\linewidth]{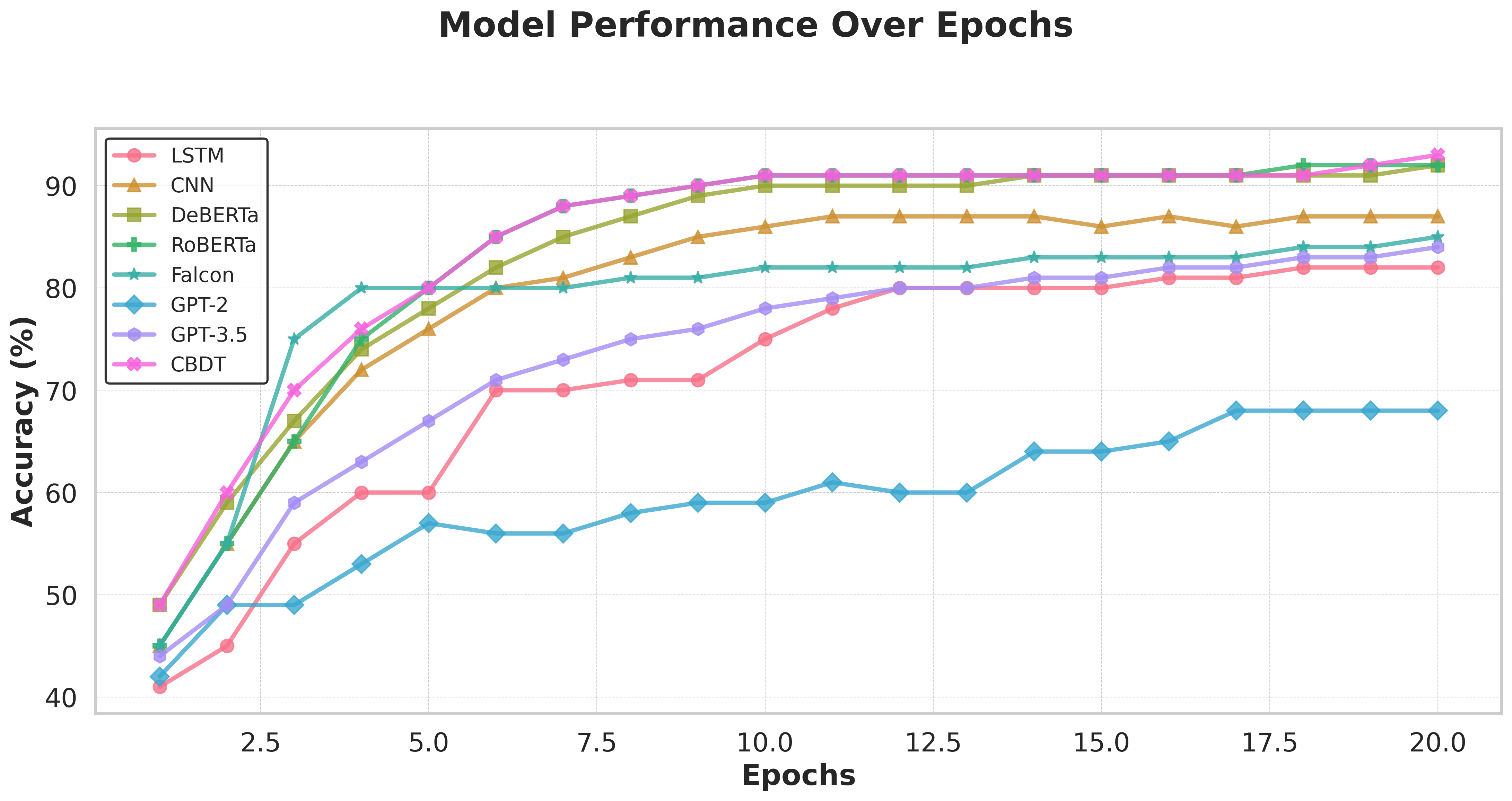}
     \caption{Performance trends of various classification models across training epochs.}
\label{fig:epochs}
\end{figure}

Figure \ref{fig:epochs} illustrates the accuracy progression of various baseline sequence classification models on the test set over multiple training epochs. The graph reflects the models' incremental learning and their improving ability to generalize as they are exposed to the training data repeatedly through each epoch.

The results, in Figure \ref{fig:epochs} for different models over 20 training epochs shows a clear trend of improvement in accuracy, with the transformer-based LMs—CBDT  {\color{green}\faLeaf} , DeBERTa and RoBERTa—demonstrating superior performance. CBDT {\color{green}\faLeaf} achieves an accuracy of about 93\%, where as DeBERTa and RoBERTa also exhibit most significant gains, starting at 49\% and 45\% accuracy, respectively, and reaching 92\% by the end of the training period. CNN shows commendable improvement, achieving 87\% accuracy, consistently maintaining this level towards the later epochs. The LSTM and Falcon models demonstrate moderate improvement, with the former reaching 82\% and the latter 85\%. GPT-2 lags behind, ending at 68\% accuracy, suggesting it is less effective for this particular task or dataset. In contrast, GPT-3.5 shows better adaptability with a final accuracy of 84\%. These results highlight the rapid advancements in model architectures, particularly transformers, and their effectiveness in learning from data.

\paragraph{Model Effectiveness: AUC Performance}

\begin{table}[h]
\centering
\caption{Area Under the Curve (AUC) values for various models, including standard deviation (SD), expressed as percentages. The highest score is highlighted in \textbf{bold}.}
\label{tab:auc_values_percentage}
\begin{tabular}{lc}
\toprule
\textbf{Model} & \textbf{AUC Value (\%) $\pm$ SD} \\
\midrule
Random Forest (RF) & 63 $\pm$ 2 \\
Support Vector Machine (SVM) & 59 $\pm$ 3 \\
Naïve Bayes (NB) & 59 $\pm$ 1.5 \\
Long Short-Term Memory (LSTM) & 63 $\pm$ 2.5 \\
Convolutional Neural Network (CNN) & 62 $\pm$ 2 \\
DeBERTa & 70 $\pm$ 1 \\
Falcon7B & 72 $\pm$ 1.2 \\
GPT-2 & 57 $\pm$ 3 \\
GPT-3.5 & 76 $\pm$ 1.5 \\
\midrule
CBDT  {\color{green}\faLeaf} (ours)  & \textbf{81 $\pm$ 1} \\
\bottomrule
\end{tabular}
\end{table}

To further quantify their performance, we compute the Area Under the Curve (AUC) metric for each model. The AUC metric is an aggregated measure of a model's ability to distinguish between classes at various classification thresholds.

Analysis of the models in Table \ref{tab:auc_values_percentage} reveals a hierarchy of performance, with the CBDT {\color{green}\faLeaf} model achieving the highest AUC at 81\%, indicative of superior accuracy and consistency. Transformer-based LMs such as DeBERTa and GPT-3.5 also show robust performance, underscoring the advantages of advanced model architectures. Conversely, traditional models like SVM and Naïve Bayes deliver modest results, while GPT-2 falls behind its newer counterparts, underscoring the rapid advancements in AI technologies. These findings emphasize the critical importance of selecting models based on both performance and reliability for classification tasks.

\subsection{Evaluating Context Transformer for Token Classification}

\begin{table}[h]
    \centering
    \caption{Comparative Performance of various models on two test sets. \textbf{Top:} Our Test Set and \textbf{bottom:} CoNLL. Higher scores are better and highlighted in \textbf{bold}. All models are fine-tuned; Falcon7B and GPT-3.5 are tested on zero-shot (ZS) and few-shot (FS) settings, respectively.}
    \begin{tabular}{llccc}
    \toprule
    & Model & Precision (\%) & Recall (\%) & F1-Score (\%)  \\
    \midrule
    {\multirow{8}{*}{\rotatebox[origin=c]{90}{Test Set}}} & LSTM & 79.78 & 81.23 & 80.50 \\
    & CNN & 86.10 & 86.90 & 86.50 \\
    & DeBERTa & 87.00 & 87.10 & 87.05 \\
    & RoBERTa & 89.10 & 90.18 & 89.64 \\
    & Falcon7B (ZS) & 86.13 & 87.39 & 86.76 \\
    & GPT-2 (FT) & 56.36 & 56.63 & 56.50 \\
    & GPT-3.5 (FS) & 89.80 & 90.70 & 90.25 \\
     \cmidrule{2-5}
    & CBDT  {\color{green}\faLeaf}  (ours) & \textbf{92.10} & \textbf{93.04} & \textbf{92.57} \\
    \midrule
    {\multirow{8}{*}{\rotatebox[origin=c]{90}{CoNLL}}}& LSTM & 77.00 & 79.00 & 77.99 \\
    &CNN & 78.00 & 81.00 & 79.47 \\
    &DeBERTa & 81.00 & 82.30 & 81.64 \\
        &RoBERTa & 83.00 & 85.00 & 83.99 \\

       &Falcon7B (ZS) & 82.00 & 81.00 & 81.50 \\
 &GPT-2 (FT) & 52.77 & 52.97 & 52.87 \\
    &GPT-3.5 (FS) & 82.30 & 84.50 & 83.39 \\
     \cmidrule{2-5}
    &CBDT  {\color{green}\faLeaf}  (ours) & \textbf{87.00} & \textbf{88.70} & \textbf{87.84} \\
    \bottomrule
    \end{tabular}
    \label{tab:tc_combined}
    \vspace{-5mm}
\end{table}

\noindent Table \ref{tab:tc_combined} shows the performance of various models in token classification tasks, allowing us to derive the following insights:

Our proposed CBDT  {\color{green}\faLeaf}  model comes as the top-performing model for token classification on the primary test set. With an F1-score of  92.57\%, it performs best, reflecting its efficiency in identifying and localizing biases at the token level within text sequences. 
The LSTM and CNN, two fundamental neural network architectures, demonstrate commendable performances with F1-scores of 80.50\% and 86.50\%, respectively, on the primary test set. This suggests that while traditional neural networks can effectively tackle token-level classification, more advanced architectures offer enhanced precision and recall.

The transformer-based models, particularly DeBERTa, GPT-3.5, and RoBERTa, establish their superiority in the token classification realm. Their deep architectures, combined with the ability to capture long-range dependencies in text, contribute to their success. Among these, GPT-3.5's few-shot approach exhibits impressive results, although it doesn't surpass the performance of the CBDT  {\color{green}\faLeaf}  model. The slight advantage of CBDT  {\color{green}\faLeaf}  over models BERT-like models might stem from its design, specifically tailored for bias detection. Compared to GPT-3.5, fine-tuned GPT-2 could not achieve higher performance, which may show the model limitations in adapting to specialized tasks without extensive pre-training.  Falcon7B's zero-shot approach on the primary test set produces an F1-score of 86.76\%, showcasing the potential of zero-shot learning in token classification tasks. However, when fine-tuning is applied as in GPT-3.5's few-shot approach, the performance further improves, underscoring the benefits of model adaptation to specific tasks.

The performance drop across most models when transitioning to the CoNLL-2003 dataset, an out-of-distribution dataset, underscores the inherent challenges of adapting to unseen data. However, the CBDT  {\color{green}\faLeaf}  model again proves its performance by achieving the highest F1-score of 87.84\%, illustrating its robust generalization capabilities \footnote{In later experiments, due to better performance of these models on our test set for both tasks, we show performance on our test set only.}.
The heatmap visualization, utilizing attention scores derived from selected samples, reveals varying intensities of bias within textual content, as given in  Appendix B. The heatmap underscores that specific terms in our example sentences markedly indicate hate speech, toxicity, or discriminatory language, as evidenced by their higher attention scores.

\paragraph{Bias Detection Across Key Dimensions}
\noindent Table \ref{tab:f1_scores} presents the macro-average F1-scores for various models across different demographic categories, namely Gender, Religion, Race, Age, Sexual Orientation, Disability, Nationality, Income level. 

\begin{table*}[ht!]
\centering
\caption{Performance of the token classification task using different models with macro-averaged F1-scores (\%) on our test set. \textbf{Bold} indicates the best value; the higher the score, the better the performance.}
\begin{tabular}{lcccccccc}
\toprule
\textbf{Model }          & \textbf{Age}  & \textbf{Gender} & \textbf{Race} & \textbf{Religion} & \textbf{Disability} & \textbf{Income-level} & \textbf{Climate} & \textbf{Wellness} \\ \midrule
LSTM            & 70.23 & 72.45   & 65.10 & 68.78     & 60.32       & 75.67         & 68.89    & 70.12     \\ 
CNN             & 75.21 & 78.01   & 70.67 & 72.88     & 65.45       & 80.10         & 72.45    & 75.32     \\ 
DeBERTa         & 78.10 & 80.34   & 72.89 & 75.67     & 68.12       & 82.45         & 75.78    & 78.09     \\ 
RoBERTa         & 80.67 & 82.90   & 75.12 & 78.01     & 70.56       & 85.23         & 78.21    & 80.45     \\ 
Falcon7B (ZS)   & 72.56 & 75.12   & 68.34 & 70.09     & 62.20       & 78.01         & 70.78    & 72.32     \\ 
GPT-2 (FT)      & 68.89 & 70.67   & 62.34 & 65.78     & 58.90       & 72.12         & 65.10    & 68.56     \\
GPT-3.5 (FS)    & 82.12 & 85.78   & 78.09 & 80.45     & 72.90       & 88.01         & 80.34    & 82.78     \\ \midrule
CBDT  {\color{green}\faLeaf} (ours)             & \textbf{85.56} & \textbf{88.10}   & \textbf{80.67} & \textbf{82.56}     & \textbf{75.01}       & \textbf{90.09 }        & \textbf{82.90 }   & \textbf{85.12}     \\ \toprule
\end{tabular}
\label{tab:f1_scores}
\vspace{-5mm}
\end{table*}

Table \ref{tab:f1_scores} shows that CBDT  {\color{green}\faLeaf}  consistently performs best, showcasing its adaptability across diverse datasets. GPT-3.5 and RoBERTa also perform quite well, emphasizing the capability of transformer architectures in intricate token classifications. However, DeBERT and CNN exhibit moderate performances with noticeable variations in categories like Age, Race, and Religion, indicating potential complexities. Falcon7B, despite its zero-shot capability, performs variably, excelling in categories like Sexual Orientation but lagging in others such as Income Level. This emphasizes that its adaptability might have boundaries. LSTM, a traditional architecture, shows low-to-medium performance reinforcing the significance of newer architectures in complex bias detection tasks.

\begin{table}[hb!]
\caption{Performance Metrics for Architectural and Ensemble Variations.}
\label{table:arch_and_ensemble_variations}
\centering
\footnotesize
\vspace{2mm}
\begin{tabular}{lcc}
\toprule
\textbf{Model Variations} & \textbf{Accuracy (\%)}& \textbf{Comp. Cost}\\
\midrule
CBDT  {\color{green}\faLeaf}  (ReLU Activation) & 95.0 & Moderate \tablefootnote{Moderate: Can be run on a standard GPU with reasonable training times.} \\
CBDT  {\color{green}\faLeaf}  (Sigmoid Activation) & 96.2 & Moderate\\
CBDT  {\color{green}\faLeaf}  (tanh Activation) & 94.8 & Moderate \\
\midrule
SVM (Linear Kernel) & 87.0 & Low \tablefootnote{Low: Can be run on a standard CPU without specialized hardware.} \\
SVM (RBF Kernel) & 89.2 & Medium \\
\midrule
RandomForest (100 Estimators) & 86.5 & Medium \\
RandomForest (200 Estimators) & 86.8 & High \tablefootnote{High: Requires multiple GPUs or advanced hardware for training within a practical time frame.} \\
\midrule
LSTM (Unidirectional) & 90.0 & Moderate \\
LSTM (Bidirectional) & 91.8 & Moderate \\
\midrule
Ensemble of Best-Performing Models & 97.0 & High \\
CBDT  {\color{green}\faLeaf}  + FastText Ensemble & 96.5 & High \\
\bottomrule
\end{tabular}

\end{table}

\subsection{Impact of the CBDT  {\color{green}\faLeaf}  Model Architectural Variations}


\noindent To assess our model's robustness, we made various architectural adjustments, such as changing activation functions and exploring ensemble methods. We selected these specific architecture and configuration combinations based on factors like relevance, computational feasibility, and expected performance impact. Table \ref{table:arch_and_ensemble_variations} displays the results of these architectural variants on accuracy measure.

Table \ref{table:arch_and_ensemble_variations} shows that ensemble methods yield the highest performance, specifically 97.0\% accuracy, although at the cost of a high computational cost. Among individual algorithms, CBDT  {\color{green}\faLeaf} with different activation functions exhibit strong performances with a balanced computational requirement. Specially, the Sigmoid activation achieves the highest accuracy at 96.2\%. Traditional ML models like SVM and Random Forest lag behind in performance, and increasing the number of estimators in Random Forest hardly improves performance while escalating computational costs. LSTMs, especially the Bidirectional variant, present a favorable trade-off between performance and computational expense, scoring a 91.8\% accuracy at moderate computational cost. Overall, the results highlight the need to carefully consider the trade-offs between performance metrics and computational demands when selecting model architectures.

\subsection{Ablation Study on Impact of Fine-Tuning and Regularization Strategies of CBDT  {\color{green}\faLeaf} }
\noindent The experiment in Table \ref{table:fine_tuning_regularization} evaluates the impact of various fine-tuning and regularization strategies on the performance of the CBDT  {\color{green}\faLeaf}  model on the combined data's testset, as measured by accuracy and computational cost. 

\begin{table}[h]
\caption{Performance Metrics for Fine-Tuning and Regularization Strategies.}
\label{table:fine_tuning_regularization}
\centering
\begin{tabular}{lcc}
\toprule
\textbf{Fine-Tuning (FT) \& Regularization}& \textbf{Accuracy (\%)}& \textbf{Comp. Cost}\\
\midrule
CBDT  {\color{green}\faLeaf}  (Baseline) & 95.0 & Moderate \\
CBDT  {\color{green}\faLeaf}  (Layer-wise FT)& 96.0 & Moderate \\
CBDT  {\color{green}\faLeaf}  (Feature Extraction FT)& 95.5 & Moderate \\
\midrule
CBDT  {\color{green}\faLeaf}  (No Dropout) & 95.0 & Moderate \\
CBDT  {\color{green}\faLeaf}  (Dropout 0.5) & 94.8 & Moderate \\
\midrule
CBDT  {\color{green}\faLeaf}  (No Weight Decay) & 95.0 & Moderate \\
CBDT  {\color{green}\faLeaf}  (Weight Decay 0.01) & 95.2 & Moderate \\
\bottomrule
\end{tabular}
\end{table}

\noindent The results  Table \ref{table:fine_tuning_regularization} showcases the impact of various \textit{fine-tuning strategies} on the CBDT  {\color{green}\faLeaf}  model's performance. The baseline CBDT  {\color{green}\faLeaf}  model achieves an accuracy of 95.0\%. When layer-wise fine-tuning is employed, there's a noticeable improvement with a 1\% increase in performance compared to the baseline, suggesting its beneficial impact. On the other hand, feature extraction fine-tuning offers a modest enhancement, with a 0.5\% increase in accuracy relative to the baseline. This indicates that while layer-wise fine-tuning is more effective, extracting features can still provide a performance boost.

\begin{table*}[ht!]
\centering
\caption{Analysis of CBDT  {\color{green}\faLeaf} 's ability to identify bias within sentences, illustrating a spectrum of detection accuracy from correct to partial identifications.}
\label{tab:manual-det}
\resizebox{\textwidth}{!}{
\begin{tabular}{p{4cm}p{1cm}p{2cm}p{6cm}}
\toprule
 \textbf{Input Sentence with Expected Bias Tokens} &  \textbf{Bias Label}&\textbf{Identified Bias Token(s)} & \textbf{Explanation of Error} \\
\midrule
Successful \lightpinkhighlight{entrepreneurs} usually have a background in \lightpinkhighlight{engineering} or \lightpinkhighlight{business}. &  Biased&entrepreneurs, engineering & Correctly identifies bias tokens, indicating an awareness of stereotypes linking success in entrepreneurship specifically to engineering or business backgrounds. \\ \hline
\lightpinkhighlight{Women} in leadership roles face more \lightpinkhighlight{scrutiny} than their \lightpinkhighlight{male} counterparts. &  Biased&Women, scrutiny & Partially identifies bias by recognizing \say{Women} and \say{scrutiny}, but misses the comparative aspect with \say{male}, reflecting a nuanced understanding of gender biases in professional settings but missing the full context. \\ \hline
Residents of \lightpinkhighlight{affluent} neighborhoods are less affected by \lightpinkhighlight{environmental pollution} than those in \lightpinkhighlight{lower-income} areas. &  Biased&environmental pollution, lower-income & Successfully identifies key bias tokens related to socioeconomic status and environmental impact, highlighting the model's capability to detect biases related to social and environmental justice. \\ \hline
Individuals with \lightpinkhighlight{physical disabilities} are often perceived as \lightpinkhighlight{less productive} in the workplace. &  Biased&physical disabilities & Partially identifies the bias, recognizing \say{physical disabilities} but not \say{less productive}, indicating a limitation in capturing full bias implications regarding productivity stereotypes. \\ \hline
Older employees are seen as \lightpinkhighlight{less adaptable} and \lightpinkhighlight{slower learners} compared to younger staff. &  Biased&less adaptable, slower learners & Correctly identifies all bias tokens, demonstrating a comprehensive understanding of ageist stereotypes within professional environments. \\
\bottomrule
\end{tabular}}
\end{table*}

The \textit{regularization techniques}' effects on the model are also evident from the table. Omitting dropout maintains the accuracy consistent with the baseline, hinting at the model's inherent robustness against overfitting. However, introducing a dropout rate of 0.5 slightly diminishes the performance, reducing accuracy by 0.2\%. In contrast, the introduction of a weight decay of 0.01 marginally enhances the model's metrics by 0.2\%, suggesting its potential as a beneficial regularization technique. Across all the strategies tested, the computational cost remains moderate, indicating that these techniques do not significantly strain computational resources.  

\subsection{Assessing CBDT  {\color{green}\faLeaf}  on Bias Detection and Interpretation}
In this study, we undertook a detailed investigation to assess the CBDT {\color{green}\faLeaf} model's proficiency in identifying and interpreting biases within textual data. Leveraging a sample from our dataset comprising 50 sentences, each intricately embedded with nuanced biases, we examined the model's performance. The expected label and bias tokens within these sentences were explicitly marked, allowing for a direct comparison with the tokens identified by the model.

Due to brevity reasons, we show some sample examples of our analysis in Table \ref{tab:manual-det} that shed light on each model's ability to identify biases accurately.
Our analysis showed a varied spectrum of detection capabilities exhibited by the CBDT {\color{green}\faLeaf}  model: it successfully pinpointed biases pertaining to gender and geographical stereotypes with high accuracy. However, it showed limited success in recognizing biases linked to xenophobia and ageism, often only identifying subjects or comparative terms rather than fully understanding the implications of bias. This exploration highlights the vital need to refine the ability of LM-based models to accurately detect and interpret the complex dimensions of bias present in language. 

\section{Discussion}
\paragraph{Theoretical Impact}

The results of our study contribute to the growing corpus of research on bias detection within NLP models. Our work not only underscores the efficacy of deep learning or language models, but also enhances our understanding and strategies for bias mitigation.

The CBDT {\color{green}\faLeaf} model's outperformance in comparison to traditional baselines reaffirms the potential of transformer-based approaches, especially when applied in a layered and judicious manner to distinct tasks—sentence-level and token-level bias detection in our case. The variation in performance metrics observed across different models further illuminates the intricate ways biases manifest within neural networks. In essence, our method—integrating two transformers for dual-level bias evaluation—acts as a conduit that fuses progressive NLP techniques and language models with the nuanced realm of societal biases.

\paragraph{Practical Impact}
\noindent From a practical perspective, our research has several implications:
Developers and practitioners can have a potent tool like the CBDT  {\color{green}\faLeaf}  model. By incorporating it into their systems, they can achieve a more balanced and unbiased content analysis, thereby improving overall user experience. Given the fluid nature of content and societal norms, models that remain static risk becoming obsolete or, perform worse. Such outdated models can perpetuate harmful biases. Regular evaluations ensure that models stay relevant, accurate, and aligned with current perspectives. 

This study also highlights the versatility and adaptability of the CBDT  {\color{green}\faLeaf}  model. Its ability to discern biases at both the sentence and token levels makes it applicable across a wide array of NLP tasks, from sentiment analysis to content filtering. This research provides a blueprint for organizations and institutions looking to audit their AI systems for fairness and inclusivity. This is especially relevant in sectors like finance, healthcare, and public services, where biased decisions can have serious ramifications. The detailed analyses presented offer a valuable resource for non-experts, facilitating a clearer understanding of how biases manifest in AI and the steps taken to counteract them. This transparency is crucial for fostering public trust in AI-driven systems.

\paragraph{Limitations and Future Directions}
\noindent While our research provides meaningful contributions, it comes with certain limitations. Firstly, our evaluation predominantly centered on English language datasets, which might not entirely reflect the biases inherent in other languages. Furthermore, our study primarily focuses on detecting textual biases, which means potential biases in other mediums such as images or audio might have been overlooked. Additionally, with the ever-evolving science of LLMs, newer developments such as LlaMa2 \cite{LLaMa, inan2023llama}, GPT-series, Gemini and their subsequent iterations \cite{chang2023survey} are worth trying but were not considered in our current research.

Looking ahead, numerous promising directions are worth further exploration. One  direction is to try the versatility of the CBDT  {\color{green}\faLeaf}  model across diverse languages and cultural contexts for a more holistic view of bias detection. Another direction is to delve into the complexities of intersectional biases \cite{devinney2022theories}, where multiple layers of bias can intertwine and interact. Furthermore, gathering real-time user feedback can help continually adapting and aligning models more closely with shifting societal norms and values \cite{raza2022machine}. Finally, it would also be worthwhile to extend this work towards debiasing the content in real time.

\section{Conclusion}
\noindent The widespread utilization of NLP models in everyday tasks highlight the importance of understanding these models' behavior. This is particularly important in relation to their handling and potential propagation of biases. Our research is centered on the detection of biased content, leveraging the capabilities of transformer-based LMs, to identify and mitigate biases in textual content. Our findings underscore the effectiveness of the CBDT  {\color{green}\faLeaf}  model when compared to a variety of baseline models. We illustrate the CBDT  {\color{green}\faLeaf}  model's proficiency in identifying biases at both the sequence and token levels. The corpus construction process is needed for FAIR data adherence, for which we provided sufficient guidelines. Furthermore, the practical implications of our discoveries extend beyond theoretical discourse, as they provide concrete solutions for developers, practitioners, and other stakeholders.

\section*{Acknowledgment}

The authors extend their gratitude to the Province of Ontario, the Government of Canada through CIFAR, and the corporate sponsors of the Vector Institute for their generous support and provision of resources essential for this research. Further details on our sponsors can be found at \href{https://www.vectorinstitute.ai/#partners}{www.vectorinstitute.ai/\#partners}.

\bibliographystyle{IEEEtran}
\bibliography{references} 

\end{document}